 \documentclass[pmlr,twocolumn,10pt]{jmlr} % W&CP article

\usepackage{booktabs}
\usepackage{siunitx}
\usepackage{multirow}
\theorembodyfont{\upshape}
\theoremheaderfont{\scshape}
\theorempostheader{:}
\theoremsep{\newline}

 \title[PEFT for HMC]{Enhancing Health Mention Classification Performance: A Study on Advancements in Parameter Efficient Tuning}
\author{%
\Name{Reem Abdel-Salam} \Email{reem855@eng.cu.edu.eg} \\
\addr{Faculty of Engineering, Computer Engineering Department, Cairo University / Egypt  \\
  CaresAI/ Australia}
\AND
\Name{Mary Adetutu Adewunmi} \Email{Mary.Adewunmi@menzies.edu.au} \\
\addr{ Menzies School of Health Research, Charles Darwin University, Northern Territory / Australia  \\ CaresAI/ Australia}
}

\begin{document}
\maketitle
\begin{abstract}
Health Mention Classification (HMC) plays a critical role in leveraging social media posts for real-time tracking and public health monitoring. Nevertheless, the process of HMC presents significant challenges due to its intricate nature, primarily stemming from the contextual aspects of health mentions, such as figurative language and descriptive terminology, rather than explicitly reflecting a personal ailment. To address this problem, we argue that clearer mentions can be achieved through conventional fine-tuning with enhanced parameters of biomedical natural language methods (NLP). In this study, we explore different techniques such as the utilisation of part-of-speech (POS) tagger information, improving on PEFT techniques, and different combinations thereof. Extensive experiments are conducted on three widely used datasets: RHDM, PHM, and Illness. The results incorporated POS tagger information, and leveraging PEFT techniques significantly improves performance in terms of F1-score compared to state-of-the-art methods across all three datasets by utilising smaller models and efficient training. Furthermore, the findings highlight the effectiveness of incorporating POS tagger information and leveraging PEFT techniques for HMC. In conclusion, the proposed methodology presents a potentially effective approach to accurately classifying health mentions in social media posts while optimising the model size and training efficiency.

\end{abstract}

%\section{Introducation}
%\label{sec:Introducation}

\section{Introduction}
\label{sec:intro}
The emergence of social media platforms has drastically changed how individuals communicate and exchange information \cite{nabiilah2025effectiveness}. Social media platforms like Twitter and Reddit have become one of the most useful resources where people discuss a variety of issues, including health and well-being. These platforms offer an extensive amount of data that can be utilised to gain insights into public health issues, identify emerging trends\cite{park2017tracking}, and improve healthcare solutions. However, the unstructured nature and large volume of social media data make collecting useful health-related information difficult. Detecting information connected to a health report is a critical step in utilising social media data sources for public health surveillance \cite{charles2015using}. One of the most important reasons is that it enables public health officials and researchers to monitor and track the spread of diseases, identify outbreaks, and assess the impact of interventions \cite{golder2015systematic}. Also, it provides a valuable tool for the early detection of adverse drug reactions and side effects. Furthermore, it facilitates the identification of public health concerns, such as misinformation. Thus, Health Mention Classification Tasks (HMC) have emerged. The goal of health mention classification (HMC) is to determine whether or not a text contains personal health mentions \cite{lamb2013separating}. 
Pre-trained Language Models (PLMs) have recently received a lot of attention and have made significant improvements in a variety of downstream Natural Language Processing (NLP) tasks \cite{min2023recent}\cite{}, including classification of texts, question answering \cite{ngo2025quality}, disease detection \cite{kumar2025natural}, predicting user engagement in health misinformation correction \cite{kuo2025predicting} and machine translation \cite{sun2022survey}. PLMs can learn syntactic, semantic, and structural linguistic information. The fine-tuning strategy with extra classifiers has been widely used to stimulate and exploit rich information in PLMs and has achieved great performance in downstream tasks to adapt the varied knowledge contained in PLMs to diverse NLP tasks \cite{tezgider2022text,wang2021practical}. Nonetheless, as PLMs grow in size, fine-tuning larger PLMs becomes increasingly challenging in most real-world applications. A significant alternative strategy is parameter-efficient fine-tuning (PEFT) \cite{houlsby2019parameter,liao2023parameter}, in which a small set of task-specific parameters is changed while the remainder of PLM's parameters are frozen. In this manner, only one generic PLM is saved or transferred, together with the changed parameters for each task. Aside from minimising memory and training costs, PEFT equals the performance of full fine-tuning while updating less than 1\% of the PLM parameters, adapts fast to new jobs without catastrophic forgetting, and frequently displays robustness in out-of-distribution evaluation. These compelling benefits have spurred significant interest in the use of PEFT. 
This study aims to examine various strategies and assess their influence on enhancing the performance of the HMC model.   We investigate the application of part-of-speech (POS) tagger data, various PEFT techniques, enhancements to PEFT techniques, and alternative combinations of these elements.   A comprehensive series of experiments is carried out on three commonly utilised datasets: RHDM, PHM, and Illness.
The rest of the paper is organized as follows: In section \ref{sec:rlwork}, we discuss the
related work, whereas in section \ref{sec:method}, we present our method for HMC.  In section \ref{sec:resultanddicuss}, we present the results analysis and discussions
of the experiments. Finally, in section \ref{sec:conc}, we provide the conclusion of the paper

\section{Related Work}
\label{sec:rlwork}
\cite{karisani2018did} described an approach known as ``WESPAD" that acts as a regularizer for HMC tasks on Twitter data. It partitioned and distorted the embedding, allowing the model to achieve generalisation capacity \cite {jiang2018identifying}, which represented the tweets using non-contextual embeddings. LSTMs were fed these embeddings.
The authors demonstrated that using LSTM before the classification layer enhanced performance over simple SVM, KNN, and decision trees. \cite{iyer2019figurative} used idiom detection attributes from an unsupervised statistical learner and fed them to a CNN-based classifier. These characteristics increased classification performance over the CNN classifier trained on pre-trained embeddings. \cite{biddle2020leveraging} worked with both non-contextual embeddings like word2vec and contextual embeddings like ELMO and BERT, as well as sentiment data. \cite{aduragba2023improving} proposed using emotional information presented in the text for fine-tuning the BERT model. The authors proposed two approaches for that. The first approach is implicit emotion incorporation, where the BERT model is first fine-tuned for emotion detection and classification. Then this model is fine-tuned on HMC tasks. The second approach is Explicit emotion incorporation, where text is fed to the BERT model and emotion model to capture better emotions expressed in social media texts. Then the representation of the two models is fused and used for the final classification task. \cite{khan2022improving} proposed using adversarial training with a contrastive loss function to train $BERT_{large}$ and $RoBERTA_{large}$ in order to improve model performance. First, the text is fed to the model. Then Fast Gradient Sign Method (FGSM) \cite{liu2019sensitivity} is used to generate adversarial examples. These adversarial examples are then fed to the model. Barlow Twins loss function is then utilised for both representations (normal model, perturb model) as a contrastive loss function \cite{zbontar2021barlow}.  Then the final loss is the weighted sum of two cross-entropy losses and a contrastive loss.
\cite{naseem2022identification} presented a
a new dataset of Reddit posts called the Reddit Health Mention Dataset (RHMD). In addition, the proposed HMCNET aims to improve HMC by combining target keyword (disease or symptom phrase) detection and user behaviour hierarchically. The target keyword module computes the literal usage of a target keyword score in a given text. The user behaviour module extracts some features related to the posts, such as user social interactions, upvotes, downvotes, and emotional features. \cite{aduragba2023improving} proposed a multi-task learning approach for HMC tasks by combining literal word meaning prediction as an auxiliary task. In order to annotate the dataset with a literal meaning label, the authors proposed to use a pseudo-literal label. If a text is either labelled health-mention or other mention, then it will be marked as literal; otherwise, it will be labelled as non-literal. Furthermore, the authors proposed a new dataset, the Nairaland health mention dataset (NHMD), a new dataset compiled from a dedicated Nigerian web forum and consisting of four prominent diseases (HIV/AIDS, malaria, stroke, and Tuberculosis)\cite{aduragba2023improving}.

\section{Materials and Methods}
\label{sec:method}
In this section, we discuss the proposed methodology and techniques. In addition, we explore various datasets used.

\subsection{Datasets}
\label{data}
Three HMC-related datasets are used in this study. These datasets are gathered from different social media platforms such as Twitter and Reddit. We mainly focus on using PHM2017 \cite{KarisaniAg18}, Illness \cite{karisani2021contextual}  and  RHMD \cite{naseem2022identification}.
These datasets have been annotated to identify mentions of health-related concepts in social media content, such as health mentions/non-health mentions or figurative mentions. Table \ref{table:data-distribution}  presents the summary of all HMC datasets.

\begin{table*}[!h]
\begin{tabular}{|c|c|c|c|}
\hline
Dataset                  & Label               & Size   & Disease                                                                                                                    \\ \hline
\multirow{4}{*}{PHM2017} & Non-health          & 1145   & \multirow{4}{*}{\begin{tabular}[c]{@{}c@{}}Alzheimer, Cancer, Depression, Heart attack, Parkinson,\\  Stroke\end{tabular}} \\ \cline{2-3}
                         & Awareness           & 2343   &                                                                                                                            \\ \cline{2-3}
                         & Other-mention       & 473    &                                                                                                                            \\ \cline{2-3}
                         & Self-mention        & 283    &                                                                                                                            \\ \hline
\multirow{3}{*}{RHDM}    & Figurative Mentions & 3,225  & \multirow{3}{*}{Non}                                                                                                       \\ \cline{2-3}
                         & Non-Health Mentions & 3,430  &                                                                                                                            \\ \cline{2-3}
                         & Health mentions     & 3,360  &                                                                                                                            \\ \hline
\multirow{2}{*}{Illness} & Positive            & 3,940  & \multirow{2}{*}{Alzheimer, Parkinson, Cancer, Diabetes}                                                                    \\ \cline{2-3}
                         & Negative            & 18,720 &                                                                                                                            \\ \hline
\end{tabular}
\caption{Label distribution for each HMC dataset
.}
\label{table:data-distribution}
\end{table*}

\subsection{Methodology}
\label{method}
Symptoms or disease-related keywords are typically used to extract social media posts connected to health mentions. People on social media, on the other hand, frequently employ slang and many renderings of a term, contributing to the high noise level of social media posts. As a result, the presence of a symptom or disease term does not always imply that it is health-related. Which makes it hard for HMC tasks to detect whether a given text is a health mention or not. To address this issue of correctly identifying a word's meaning given its context. Most state-of-the-art techniques try to model context by either incorporating more features, such as emojis and sentiment, or using contrastive adversarial learning \cite{aduragba2023improving,khan2022improving}.
In this work, we address this problem from a different perspective.  The problem of the hardness of the HMC task is due to the domain shift between pre-trained transformer models and the new domain of health mentions, which might have some wording related to the medical domain or disease. In addition to the domain shift, the a lack of context understanding in terms of part-of-speech tagging and dependency parsing. For instance, the dependency of the disease may determine whether its health is mentioned or not, if it refers to a person's condition. If a model could correctly identify this relationship performance might improve. To address these issues we employ multiple experiments
\begin{itemize}
   \item Incorporating POS information
  \item Masked Language modelling on the target domain, followed by fine-tuning or PEFT
  \item Fine-tuning pre-trained models or using PEFT techniques directly.
  \item Introducing some modifications on PEFT techniques.
\end{itemize}
In this section, approaches are discussed briefly.
\subsubsection{Incorporating POS information}
Part of speech tagging (POS) is important to determine the relation between the words and their dependencies. Sometimes to understand the true meaning of the word, and to resolve word sense disambiguation, one needs to determine the POS category for a given word. From this perspective, we aim to investigate the impact of adding such information to language models. We consider two approaches to
incorporate :

\begin{itemize}
\item Intermediate Task Fine-tuning Approach \cite{chang2021rethinking}: In this approach, we used a fine-tuned model on the POS task, and we fine-tuned it again on the new downstream task (HMC in our case). The idea behind intermediate fine-tuning is that if both activities are connected, the language knowledge acquired in the intermediate work can help with understanding the target task.
\item Representation Fusion Approach \cite{wang2020multi}: In this technique, we leverage both representations from the original pre-trained model and the fine-tuned model on POS tasks. First text is fed to both models; then we fuse the output representation of both models; and finally, the fused representation is fed to the classifier head. 

\end{itemize}

\subsubsection{Masked Language Modeling}
Domain shift is a common issue in most of the NLP tasks \cite{wang2019exploring}, where the pre-trained model is originally trained on different domains and distributions from the downstream task problems. As a result of the variation in the new distribution of training sets, the model fails to utilise the nature of the new information being trained on and exhibits poor generalisation. One way to overcome this is by pre-training the model on the new target domain. Then fine-tune the new model on the target task and domain. Another way is to make use of prompt-tuning, as well as prefix-tuning. In our case, we explore the impact of performing pre-training first, followed by either fine-tuning or one of the PEFT techniques. A major issue presented is the lack of a new dataset to pre-train the model on. In order to overcome this, we utilise the dataset. If the HMC task is on the RHMD dataset, then we pre-train the model using the Illness and PHM2017 datasets. If the HMC task is carried out using Illness dataset then we pre-train the model using the RHMD and PHM2017 datasets.

\subsubsection{Parameter Efficient fine-tuning}
Fine-tuning pre-trained models has been widely demonstrated to be useful in a variety of NLP tasks. However, fine-tuning the entire model is inefficient because it always results in an entirely new model for each task.
Many current research studies propose just fine-tuning a small percentage of the parameters while keeping the majority of the parameters shared across multiple projects. These methods function remarkably well and have been proven to be more stable than their fully fine-tuned counterparts. 
These methods are widely known parameter-efficient tuning methods (PEFT), and it includes P-tuning v2 \cite{liu2021p}, Prefix-tuning \cite{li2021prefix}, prompt-tuning \cite{lester2021power}, soft-prompting \cite{liu2023gpt} and LoRa \cite{hu2021lora}.
%In prompt-tuning.
\paragraph{Prompt-tuning}
This entails introducing a specific prompt or instruction to the input data prior to feeding the input data into the pre-trained language model.
This prompt can take the form of a word, phrase, or sentence that directs the language model to produce a particular output. As an illustration, the prompt in a text classification job may be a label or a category that the language model has been trained to anticipate. A prompt can make it possible for the language model to produce outputs that are more accurate.
Let V represent the language model's vocabulary, and e represent the language model's embedding layer. When prompting is discrete, the prompt tokens \textbf{``It," ``feels," and ``[MASK]"} belong to V and can be used to categorise a sentiment. For instance, the input embedding sequence is written as \textbf{[e(x), e(``It"), e(``feels"), e(``[MASK]")]} given the input text \textbf{x=\``happy bday!"}.

\begin{figure}

\includegraphics[width=\linewidth]{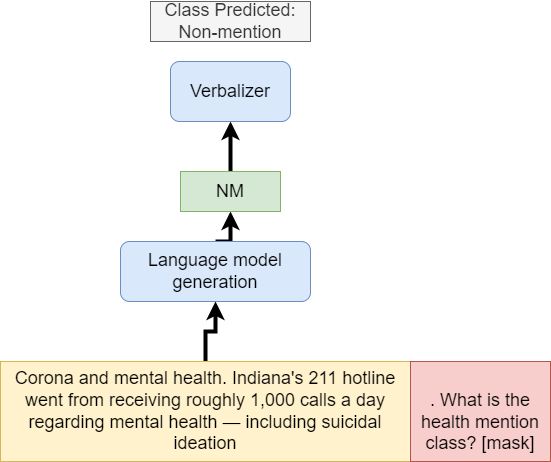}
\caption{Prompt architecture for the RHMD dataset.}
\label{pormpt}
\end{figure}

In HMC tasks, different prompts and verbalizers (target class) are explored, according to the target dataset.
In RHMD2022, we have used the as \textbf{``. What is health mention class? [mask]."}, as for the verbalizers, two sets have been used as our category to predict \textbf{``FM, NM, HM"}, which refers to figurative mention, non-health mention, and health mention. The second set was \textbf{``Figure, Non, Health"} which refers to figurative mention, non-health mention, and health mention. The architecture can be shown in figure \ref{pormpt}
In the Illness dataset, two different prompts have been used: \textbf{``. Does this text indicate a health mention of {name of the disease}? [mask]."} and \textbf{`` . Is a person diagnosed with {name of the disease}? [mask]."} . For the verbalizers two sets have been explored \textbf{``Yes, No"}, and \textbf{``Positive, Negative"}.
In the PHM2017 dataset, two different prompts have been used: \textbf{``. What is health mention class?  [mask]."} and \textbf{`` . What is health mention type for this  {name of the disease} disease? [mask]."} . As for the verbalizers, we have used one set \textbf{``NM, "A" "OM" and "SM"} which refer to non-health mention, awareness, other mention, and self mention.

\begin{figure}[htbp]

\includegraphics[width=\linewidth]{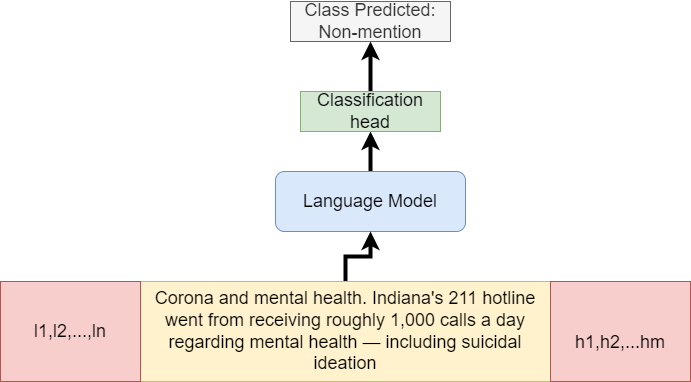}

\caption{Modified Soft-prompting architecture to include virtual tokens before and after the text.}
\label{pre-post-soft-prompt}
\end{figure}

\paragraph{Soft-Prompting}
Instead of adding a hard prompt, as in prompt-tuning, a soft prompt involves adding virtual learnable tokens (continuous prompts) with the input text.
The input embedding sequence is expressed as \textbf{[ h0,..., hi, e(x)]}, given the trainable continuous embeddings \textbf{[h0,..., hi]} representing virtual tokens.
We propose two modifications. The first modification is to add virtual token wrapping text: The input embedding sequence is expressed as \textbf{[ h0,..., hi, e(x), hi+1,....,hn]}, as shown in figure \ref{pre-post-soft-prompt}

\begin{figure}[htbp]

\includegraphics[width=\linewidth]{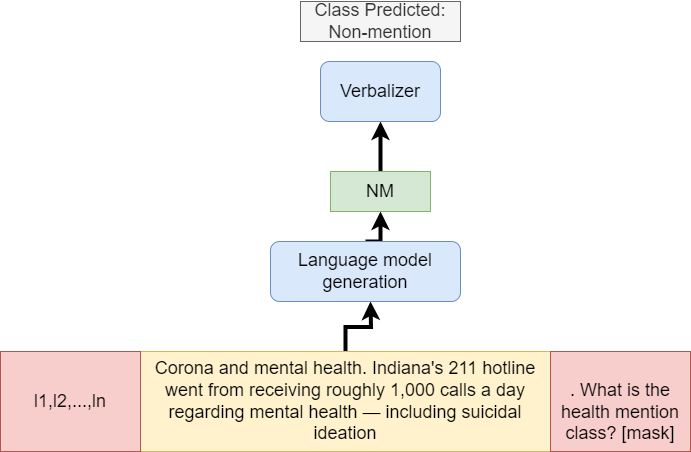}

\caption{Modified Soft-prompting architecture to include a virtual token before the text and a hard prompt after the text.}
\label{prompting-soft-prompt}
\end{figure}

The second modification is to combine both prompt-tuning with soft prompting as shown in figure \ref{prompting-soft-prompt}, so that the input sequence could be represented as 
\textbf{[h0,..., hi, e(x), e(prompt), e(``[MASK]")] } given an input text x  and a prompt.

\paragraph{P-tuning V2/Prefix-tuning}
P-tuning V2 entails adding trainable tensors to each transformer block rather than just the input embeddings as in soft-prompting. This allows more capacity and generalisation for each layer while being efficient in training.

\paragraph{P-tuning V2 / Prefix-tuning with LoRa}
In this approach, P-tuning V2 is combined with LoRa to allow better generalisation and performance.

\section{Results and Discussion}
\label{sec:resultanddicuss}
\subsection{Experimental Setup}
All models were trained using a batch size of 8 using the AdamW \cite{loshchilov2018fixing} optimiser, with a weight decay of 0.001. A cosine annealing learning rate scheduler was utilised with a maximum learning rate of $3e-5$ and a minimum learning rate of $1e-8$. All models are tuned  for
5 epochs, and the best model performing on the validation set is saved based on the F1-micro score. We created our own splits because the dataset providers did not supply them. To produce the train, and test sets for each dataset, we perform a random 70\%/30\%/ split. We utilised the same experimental parameters and 5-fold cross-validation for all models to ensure consistency.
F1-score was used as an evaluation metric.
For the RHMD dataset, the first 6 layers are kept frozen, whilst, in the Illness dataset, the first 2 layers are kept frozen in the PEFT setting. For, the PHM dataset, only embedding is kept frozen in the PEFT setting.
\subsection{Results}
Table \ref{table:allapproachs} presents the results of the proposed framework in terms of the F1-micro score for the test set. For most of the dataset, RoBERTa-based models outperform BERT-based models.
The PHM2017 datasets come in first place in prompt-tuning with a 91.5 F1-score, while in second place representation fusion approach POS. It is worth noting that soft-prompting with RoBERTa achieves the lowest results compared to other techniques and BERT-based techniques.
For the RHMD dataset, the top performing model is intermediate POS with prompt-tuning with an  82.2 F1-score, while soft prompting with hard prompting with an F1-score of 81.7 comes in second place. This shows the impact of using POS as well as prompt-tuning to improve PLM representation of word context to correctly perform HMC tasks. For the Illness dataset, the top-performing model was prompt-tuning with an F1-score of 95.5, followed by the representation fusion approach POS with an F1-score of 95.3.

Soft-prompting with it is proposed in new versions, and soft-prompting with hard prompts outbeats normal soft-prompting with a high margin of 2\% or more for RoBERTa-based models. However, the performance slightly decreased when compared to prompt-tuning in PHM2017 and the Illness datasets. However, performance increases with the RHMD dataset. Pre-training and then fine-tuning do not seem to improve performance drastically. For POS incorporation, the performance of PLM improves with the Representation Fusion Approach POS for PHM2017 and the ILLness dataset. Furthermore, for the RHMD dataset, Intermediate POS with prompt tuning performs better than the other variants of POS incorporation.

Table \ref{table:allapproachs} gives an insights about domain adaptation,
\begin{itemize}
\item Single-Source (In-Domain) -> Single-Target (In-Domain), where the model is trained and evaluated on a target dataset.
\item Multiple Source (In Domain) - Single Target (Out Domain): where in this setting we pre-train the model against other two datasets, the slightly fine-tuned model on the target dataset. 

\end{itemize}
It can be concluded that multiple sources (in domain) - single target (out domain) achieves comparable similar results when compared to
single-source (in-domain) -> single-target (in-domain), with a performance drop of around of 0.5\%.

Overall, the results demonstrate the feasibility and generalisation of our proposed approach. 
Table \ref{table:verb-illness-prompt}, investigates the performance of different verbalizers along with different prompts of the model performance. ``Yes, NO" verbalizers outperformed ``Positive and Negative by 0.5\%. Further, mentioning task problems in prompt questions improves performance by 0.1\%.
Similarly, Table \ref{table:verb-RHMD-prompt}, investigates the performance of different verbalizers for the RHMD dataset. The model was able to perform better with keywords near to original labels, such as Figure for Figurative class, Non for the Non-Health mention, and Health for the Health mention.
Finally, Table \ref{table:verb-PHM-prompt}  indicates similar results, as table \ref{table:verb-illness-prompt} mentions that the task problem in the prompt question improves performance in this case by 1\%.

% \usepackage{multirow}

% \usepackage{multirow}
\begin{table*}[]
\begin{tabular}{|c|c|c|c|}
\hline
\multirow{2}{*}{Model}   & \multirow{2}{*}{Prompt}                                                                                                         & \multirow{2}{*}{Verbalizer} & \multirow{2}{*}{F1-Score} \\
                         &                                                                                                                                 &                             &                           \\ \hline
\multirow{3}{*}{RoBERTA} & \multirow{2}{*}{``Is a person diagnosed with a disease type? [MASK]."}                                                          & Yes, No                     & 95.4                      \\ \cline{3-4} 
                         &                                                                                                                                 & Positive, Negative          & 95                        \\ \cline{2-4} 
                         & ``Does this text indicate a health mention of disease type? [MASK]." & Yes, No                     & 95.5                      \\ \hline
\end{tabular}
\caption{Investigating the effect of different prompts and verbalizers on Illness dataset}
\label{table:verb-illness-prompt}
\end{table*}

% \usepackage{multirow}

% \usepackage{multirow}
\begin{table*}[]
\begin{tabular}{|l|l|l|l|}
\hline
\multirow{2}{*}{Model}   & \multirow{2}{*}{Prompt}                                                                & \multirow{2}{*}{Verbalizer} & \multirow{2}{*}{F1-Score} \\
                         &                                                                                        &                             &                           \\ \hline
\multirow{2}{*}{RoBERTA} & \multirow{2}{*}{`` What is the health mention class? [MASK]."} & FM, NM, HM                  & 80.8                      \\ \cline{3-4} 
                         &                                                                                        & Figure, Non, Health         & 80.9                      \\ \hline
\end{tabular}
\caption{Investigating effect of different verbalizers on RHMD dataset}
\label{table:verb-RHMD-prompt}
\end{table*}
% \usepackage{multirow}
\begin{table*}[]
\begin{tabular}{|l|l|l|l|}
\hline
\multirow{2}{*}{Model}   & \multirow{2}{*}{Prompt}                                                          & \multirow{2}{*}{Verbalizer} & \multirow{2}{*}{F1-Score} \\
                         &                                                                                  &                             &                           \\ \hline
\multirow{2}{*}{RoBERTA} & ``What is health mention type for this  disease? [MASK]". & NH,A,OM,SM                  & 91.5                      \\ \cline{2-4} 
                         & ``What is the health mention class? [MASK]."            & NH,A, OM,SM                  & 90.8                      \\ \hline
\end{tabular}
\caption{Investigating the effect of different verbalizers on PHM2017 dataset}
\label{table:verb-PHM-prompt}
\end{table*}

% \usepackage{multirow}
% \usepackage{multirow}
\begin{table*}[]
% \usepackage{multirow}
% \usepackage{multirow}
\begin{tabular}{|c|c|ccc|}
\hline
\multirow{2}{*}{Model}    & \multirow{2}{*}{Technique}                                     & \multicolumn{3}{c|}{Dataset}                                       \\ \cline{3-5} 
                          &                                                                & \multicolumn{1}{c|}{PHM2017} & \multicolumn{1}{c|}{RHMD} & Illness \\ \hline
\multirow{10}{*}{RoBERTa} & Prefix-tuning                                                  & \multicolumn{1}{c|}{90.8}    & \multicolumn{1}{c|}{78.6} & 94.5    \\ \cline{2-5} 
                          & Prompt-tuning                                                  & \multicolumn{1}{c|}{91.5}    & \multicolumn{1}{c|}{80.9} & 95.5    \\ \cline{2-5} 
                          & Soft-prompting                                                 & \multicolumn{1}{c|}{70.6}    & \multicolumn{1}{c|}{79.4} & 94.7    \\ \cline{2-5} 
                          & Soft-prompting with Hard prompting                             & \multicolumn{1}{c|}{91}      & \multicolumn{1}{c|}{81.7} & 95.1    \\ \cline{2-5} 
                          & Intermediate POS with prefix-tuning                            & \multicolumn{1}{c|}{90.9}    & \multicolumn{1}{c|}{80.7} & 88.6    \\ \cline{2-5} 
                          & Intermediate POS with Prompt-tuning                            & \multicolumn{1}{c|}{90.4}    & \multicolumn{1}{c|}{82.2} & 94.3    \\ \cline{2-5} 
                          & Representation Fusion Approach POS                             & \multicolumn{1}{c|}{91.1}    & \multicolumn{1}{c|}{81.2} & 95.3    \\ \cline{2-5} 
                          & pre-training then fine-tuning                                  & \multicolumn{1}{c|}{-}       & \multicolumn{1}{c|}{81.1} & 95      \\ \cline{2-5} 
                          & Pre-training then prefix-tuning                                & \multicolumn{1}{c|}{-}       & \multicolumn{1}{c|}{78.6} & 95      \\ \cline{2-5} 
                          & \begin{tabular}[c]{@{}c@{}}Baseline\\ Fine-tuning\end{tabular} & \multicolumn{1}{c|}{90}      & \multicolumn{1}{c|}{80}   & 91      \\ \hline
\multirow{5}{*}{BERT}     & Prefix-tuning                                                  & \multicolumn{1}{c|}{90.7}    & \multicolumn{1}{c|}{79.8} & 94.2    \\ \cline{2-5} 
                          & Soft-prompting                                                 & \multicolumn{1}{c|}{90.8}    & \multicolumn{1}{c|}{78.3} & 94.2    \\ \cline{2-5} 
                          & Soft-prompting v2                                              & \multicolumn{1}{c|}{90.6}    & \multicolumn{1}{c|}{79.2} & 95.1    \\ \cline{2-5} 
                          & Prefix-tuning with LoRa                                        & \multicolumn{1}{c|}{90.0}    & \multicolumn{1}{c|}{81.5} & 94.8    \\ \cline{2-5} 
                          & \begin{tabular}[c]{@{}c@{}}Baseline\\ Fine-tuning\end{tabular} & \multicolumn{1}{c|}{90}      & \multicolumn{1}{c|}{79}   & 91      \\ \hline
\end{tabular}
\caption{Main Results Between PLMs Fine-Tuning Baselines and Our Proposed Framework.}
\label{table:allapproachs}
\end{table*}

\section{Discussion}

In this work, We exploit different 
strategies to improve model performance in HMC tasks. First, we explore the idea of out-of-domain distribution and how to overcome it by using masked language modelling tasks, followed by fine-tuning. However, such a method is costly in training. Hence, we explore the idea of using various PEFT techniques for HMC tasks. In addition, we propose different combinations of strategies to train with. Finally, we explore the idea of adding POS information to the system and how it impacts it positively and negatively. Our proposed techniques outperform both  BERT and RoBERTa baseline methods on three public datasets. The main contributions of this paper are: 
\begin{enumerate}
  \item In this work, extensive experiments are carried out, ranging from masked language modelling (MLM) to account for domain adaptation to using PEFT for efficient training.
  \item We show that incorporating language models that account for grammar can help improve performance by a magnitude ranging from 1\% to 3\%.
  \item We show and propose that merging different techniques, such as soft-prompting with prompt-tuning,  and adding soft prompts or prefixes at the beginning of the text and at the end, stabilises training. 
\end{enumerate}

\section{Conclusion}
\label{sec:conc}
In this paper, we have presented different techniques for HMC tasks, including PEFT techniques and different combinations thereof, such as Soft-prompting with hard prompting and prefix-tuning with LoRa. In addition, we propose the usage of POS information presented in the text in two modalities: the first modality uses an intermediate representation from POS models, and the other modality uses a representation fusion approach for POS. Furthermore, we have investigated domain shift and generalisation using two approaches.
Extensive experiments have been carried out across three datasets to study the impact of each technique.  Results suggest that the proposed techniques outperform the baseline.

\bibliography{jmlr-sample}

%\appendix

%\section{First Appendix}\label{apd:first}

%This is the first appendix.

%\section{Second Appendix}\label{apd:second}

%This is the second appendix.

\end{document}